\documentclass[9.9pt,logo,copyright]{nvidiatechreport}
\usepackage{amsmath}
\usepackage{amssymb}
\usepackage{epsfig}
\usepackage{lipsum}
\usepackage{amsthm}
\usepackage{comment}
\usepackage{multirow}
\usepackage{subcaption}
\usepackage{xspace}
\usepackage{setspace}
\usepackage[utf8]{inputenc} %
\usepackage[T1]{fontenc}    %
\usepackage{enumitem}
\usepackage{makecell}
\usepackage{gensymb}
\usepackage{array}
\usepackage{nimbusmononarrow}

\usepackage{parskip}        %
\usepackage{url}            %
\usepackage{booktabs}       %
\usepackage{amsfonts}       %
\usepackage{nicefrac}       %
\usepackage{microtype}      %
\usepackage{booktabs}
\usepackage{multirow}
\usepackage[authoryear,sort&compress,round]{natbib}

\def\vs{\emph{vs}.\xspace}

\def\ie{i.e.\xspace}

\usepackage[nameinlink]{cleveref}
\crefname{equation}{Eq.}{Eqs.}
\crefname{figure}{Fig.}{Figs.}
\crefname{section}{Sec.}{Sec.}
\crefname{appendix}{App.}{App.}
\crefname{table}{Tab.}{Tabs.}
\crefname{algorithm}{Algo}{Algo}
\crefname{thm}{Thm}{Thm}
\Crefname{thm}{Thm}{Thm}
\crefname{prop}{Prop}{Prop}

\newcommand{\crefnames}[3]{%
  \@for\next:=#1\do{%
    \expandafter\crefname\expandafter{\next}{#2}{#3}%
  }%
}

\title{Edify 3D: Scalable High-Quality 3D Asset Generation}

\author{
    NVIDIA\footnote{A detailed list of contributors and acknowledgments can be found in~\cref{sec:contributors} of this paper.}
}

\begin{abstract}
    
We introduce Edify 3D, an advanced solution designed for high-quality 3D asset generation.
Our method first synthesizes RGB and surface normal images of the described object at multiple viewpoints using a diffusion model.
The multi-view observations are then used to reconstruct the shape, texture, and PBR materials of the object.
Our method can generate high-quality 3D assets with detailed geometry, clean shape topologies, high-resolution textures, and materials within 2 minutes of runtime.

\end{abstract}

\begin{document}
\maketitle

\begin{figure}[h]
    \centering
    \includegraphics[width=\linewidth,page=1]{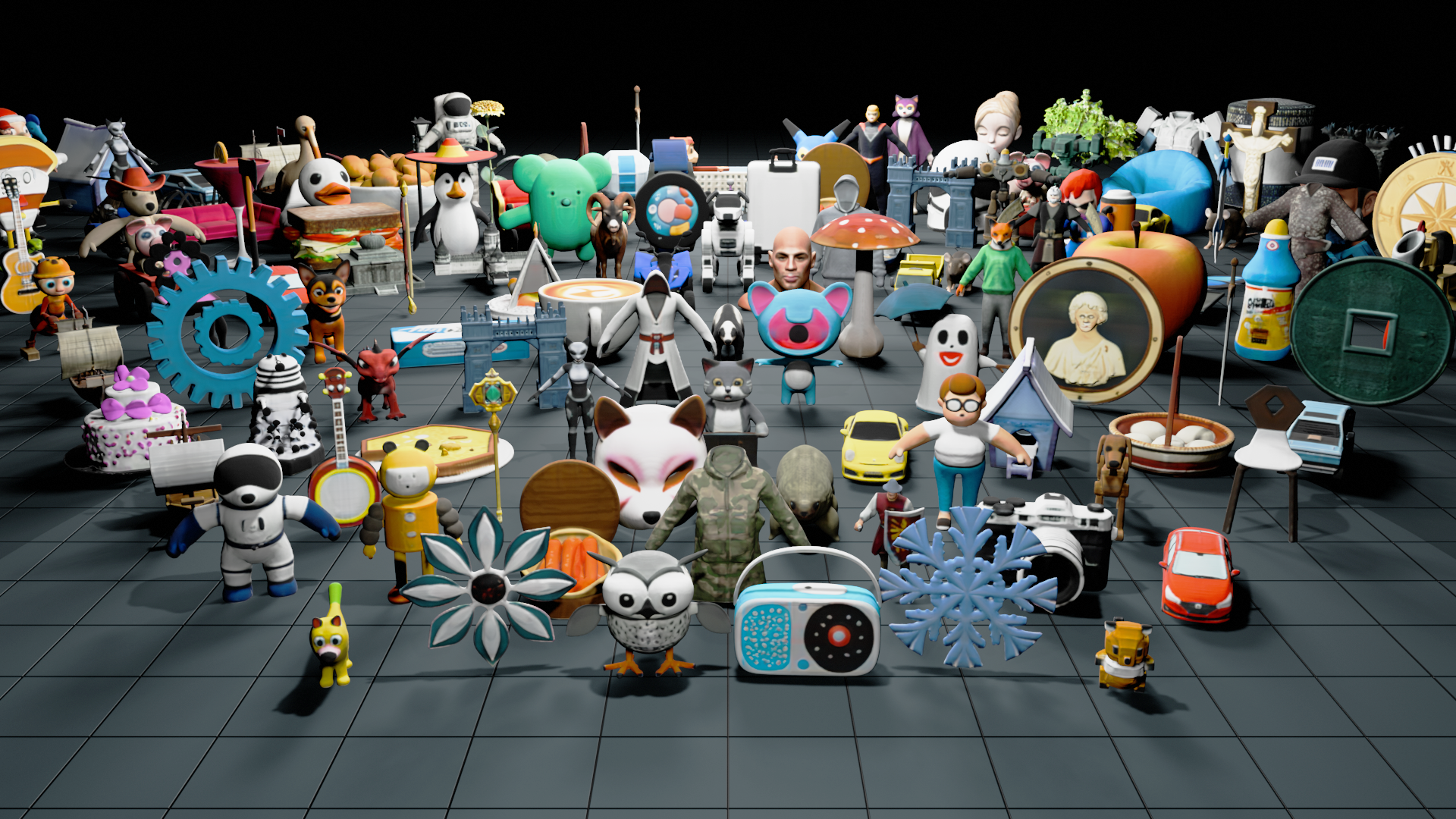}
    \caption{
        Edify 3D is a model designed for high-quality 3D asset generation.
        With input text prompts and/or a reference image, our model can generate a wide range of detailed 3D assets, supporting applications such as video game design, extended reality, simulation, and more.
    }
    \label{fig:teaser}
\end{figure}

\abscontent

\begin{center}
    \noindent\href{https://research.nvidia.com/labs/dir/edify-3d}{\textbf{https://research.nvidia.com/labs/dir/edify-3d}}
\end{center}

\section{Introduction}
\label{sec:intro}

The creation of detailed digital 3D assets is essential for developing scenes, characters, and environments across various digital domains.
This capability is invaluable to industries such as video game design, extended reality, film production, and simulation.
For 3D content to be production-ready, it must meet industry standards, including precise mesh structures, high-resolution textures, and material maps.
Consequently, producing such high-quality 3D content is often an exceedingly complex and time-intensive process.
As demand for 3D digital experiences grows, the need for efficient, scalable solutions in 3D asset creation becomes increasingly crucial.

Recently, many research works have investigated into training AI models for 3D asset generation~\citep{lin2023magic3d}.
A significant challenge, however, is the limited availability of 3D assets suitable for model training.
Creating 3D content requires specialized skills and expertise, making such assets much scarcer than other visual media like images and videos.
This scarcity raises a key research question of how to design scalable models to generate high-quality 3D assets from such data efficiently.

Edify 3D is an advanced solution designed for high-quality 3D asset generation, addressing the above challenges while meeting industry standards.
Our model generates high-quality 3D assets in under 2 minutes, providing detailed geometry, clean shape topologies, organized UV maps, textures up to 4K resolution, and physically-based rendering (PBR) materials.
Compared to other text-to-3D approaches, Edify 3D consistently produces superior 3D shapes and textures, with notable improvements in both efficiency and scalability.
This technical report provides a detailed description of Edify 3D.

\begin{figure}[tb!]
    \centering
    \includegraphics[width=\linewidth,page=1]{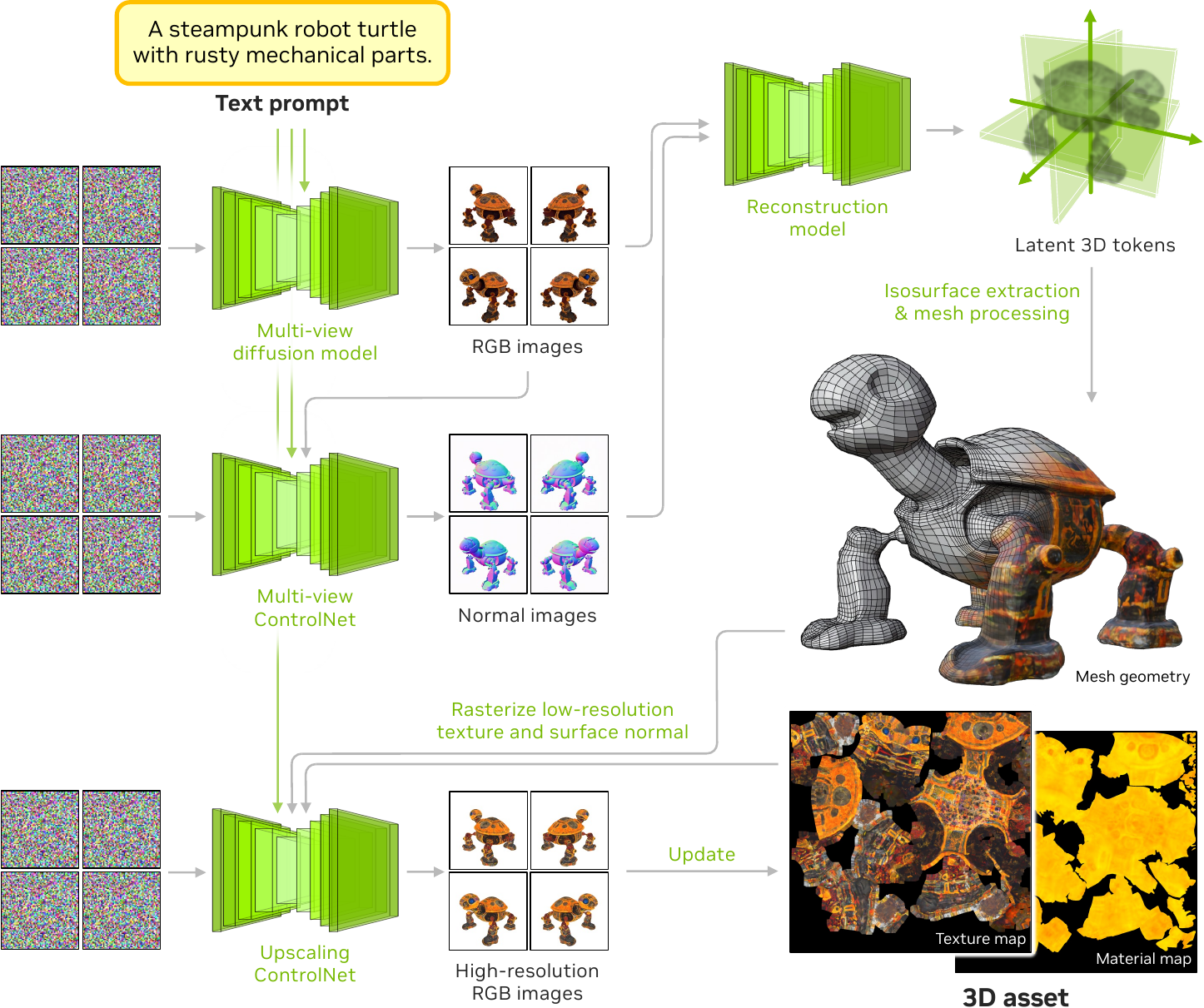}
    \caption{
        \textbf{Pipeline of Edify 3D.}
        Given a text description, a multi-view diffusion model synthesizes the RGB appearance of the described object.
        The generated multi-view RGB images are then used as a condition to synthesize surface normals using a multi-view ControlNet~\citep{zhang2023adding}.
        Next, a reconstruction model takes the multi-view RGB and normal images as input and predicts the neural 3D representation using a set of latent tokens.
        This is followed by isosurface extraction and subsequent mesh post-processing to obtain the mesh geometry.
        An upscaling ControlNet is used to increase the texture resolution, conditioning on mesh rasterizations to generate high-resolution multi-view RGB images, which are then back-projected onto the texture map.
    }
    \label{fig:diagram}
\end{figure}

\vspace{4pt}
\noindent\textbf{Core capabilities.}
Edify 3D features the following capabilities:
\begin{itemize}[leftmargin=18pt]
    \setlength\itemsep{0pt}
    \item \textbf{Text-to-3D generation}.
    Given an input text description, Edify 3D generates a digital 3D asset with the aforementioned properties.
    \item \textbf{Image-to-3D generation}.
    Edify 3D can also create a 3D asset from a reference image of the object, automatically identifying the foreground object in the image.
\end{itemize}

\vspace{4pt}
\noindent\textbf{Model design.}
The core technology of Edify 3D relies on two types of neural networks: diffusion models~\citep{song2019generative,ho2020denoising} and Transformers~\citep{vaswani2017attention}.
Both architectures have demonstrated great scalability and success in improving generation quality as more training data becomes available.
Following~\citet{li2023instant3d}, we train the following models:
\begin{itemize}[leftmargin=18pt]
    \setlength\itemsep{0pt}
    \item \textbf{Multi-view diffusion models}.
    We train multiple diffusion models to synthesize the RGB appearance and surface normals of an object from multiple viewpoints~\citep{shi2023mvdream}.
The input can be a text prompt, a reference image, or both.
    \item \textbf{Reconstruction model}.
    Using the synthesized multi-view RGB and surface normal images, a reconstruction model predicts the geometry, texture, and materials of the 3D shape.
We employ a Transformer-based model~\citep{hong2023lrm} to predict a neural representation of the 3D object as latent tokens, followed by isosurface extraction and mesh processing.
\end{itemize}
The final output of Edify 3D is a 3D asset that includes the mesh geometry, texture map, and material map.~\cref{fig:diagram} illustrates the overall pipeline of Edify 3D.

In this report, we provide:
\begin{itemize}[leftmargin=18pt]
    \setlength\itemsep{0pt}
    \item A detailed discussion of the design choices in the Edify 3D pipeline.
    \item An analysis of the scaling behaviors of model components and properties.
    \item An application of Edify 3D to scalable 3D scene generation from input text prompts.
\end{itemize}

\section{Multi-View Diffusion Model}
\label{sec:mvdiff}

\begin{figure}
    \centering
    \includegraphics[width=0.8\textwidth]{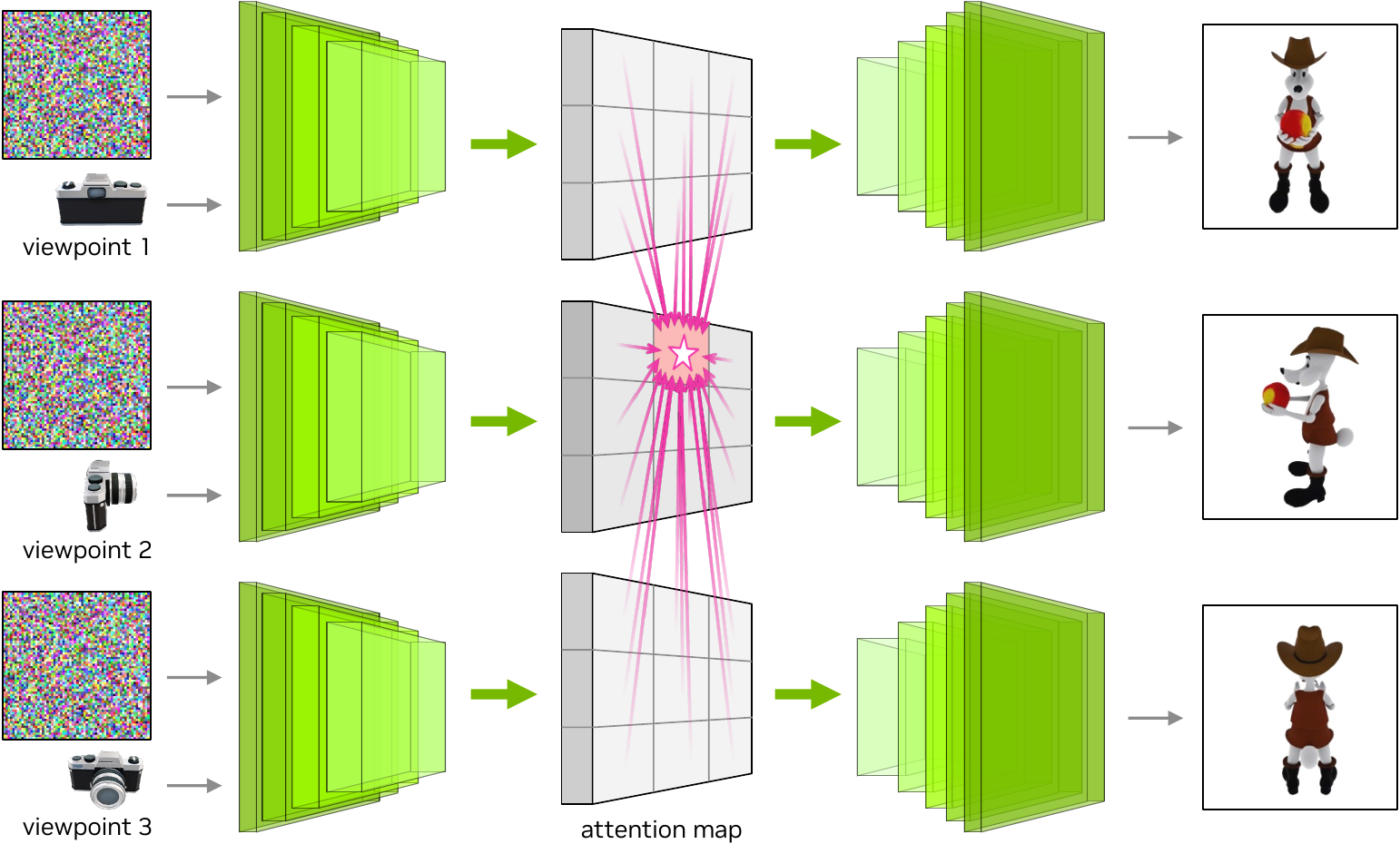}
    \caption{
        \textbf{Cross-view attention.}
        In standard diffusion models, each view is synthesized by the diffusion model independently.
        We extend the self-attention layer (yellow boxes) in our multi-view diffusion models to attend across other viewpoints using the same weights.
    }
    \label{fig:mv-attention}
\end{figure}

The process of creating multi-view images is similar to the design of video generation~\citep{chen2024v3d,videoworldsimulators2024}.
We finetune text-to-image models into pose-aware multi-view diffusion models by conditioning them with camera poses.
The models take a text prompt and camera poses as input and synthesize the object's appearance from different viewpoints.
We train the following models:
\begin{enumerate}[leftmargin=18pt]
    \setlength\itemsep{0pt}
    \item A base multi-view diffusion model that synthesizes the RGB appearance conditioned on the input text prompt as well as the camera poses.
    \item A multi-view ControlNet~\citep{zhang2023adding} model that synthesizes the object's surface normals, conditioned on both the multi-view RGB synthesis and the text prompt.
    \item A multi-view upscaling ControlNet that super-resolves the multi-view RGB images to a higher resolution, conditioned on the rasterized texture and surface normals of a given 3D mesh.
\end{enumerate}

We use the Edify Image~\citep{nvidia2024edifyimage} model as the base diffusion model architecture with a U-Net~\citep{ronneberger2015u} with 2.7 billion parameters, operating the diffusion in the pixel space.
The ControlNet encoders are initialized using the weights from the U-Net.
We extend the self-attention layer in the original text-to-image diffusion model with a new mechanism to attend across different views (\cref{fig:mv-attention}), acting as a video diffusion model with the same weights.
The camera poses (rotation and translation) are encoded through a lightweight MLP, which are subsequently added as the time embeddings to the video diffusion model architecture.

\vspace{8pt}

\noindent\textbf{Training.}
We finetune the text-to-image models on renderings of 3D objects.
During training, we jointly train on natural 2D images as well as 3D object renderings with randomly chosen numbers of views (1, 4, and 8).
The diffusion models are trained using the $\mathbf{x}_0$ parametrization for the loss, consistent with the approach used in base model training.
For multi-view ControlNets, we train the base model with multi-view surface normal images first.
Subsequently, we add a ControlNet encoder taking RGB images as input and train it while freezing the base model.

\begin{figure}[h!]
    \centering
    \small

    \begin{subfigure}[b]{0.44\textwidth}
        \centering
        \includegraphics[width=\textwidth]{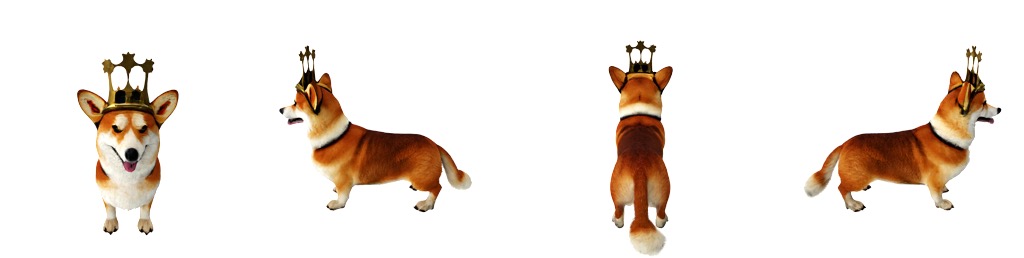}
    \end{subfigure}
    \hspace{36pt}
    \begin{subfigure}[b]{0.44\textwidth}
        \centering
        \includegraphics[width=\textwidth]{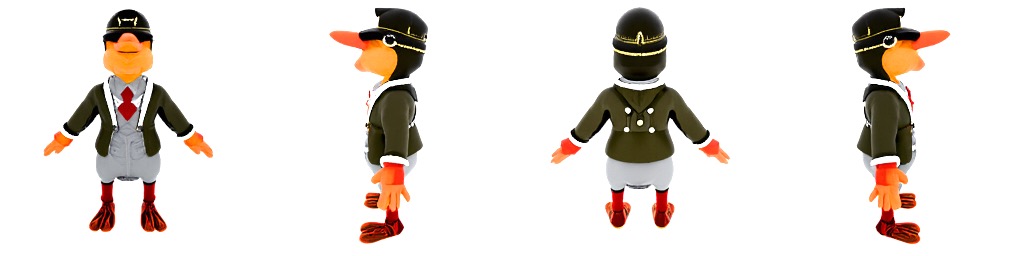}
    \end{subfigure}

    \centering
    Sampled 4 views

    \vspace{4pt}

    \begin{subfigure}[b]{0.44\textwidth}
        \centering
        \includegraphics[width=\textwidth]{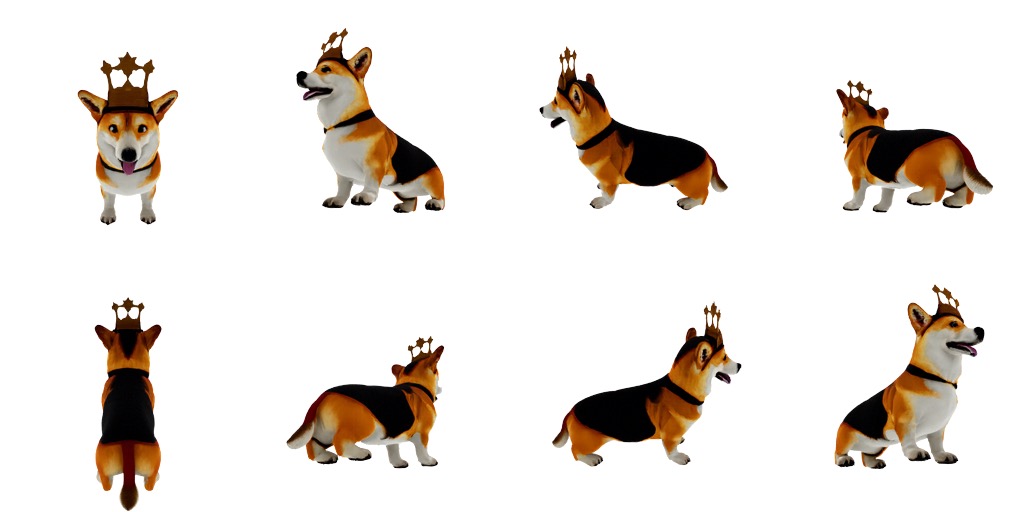}
    \end{subfigure}
    \hspace{36pt}
    \begin{subfigure}[b]{0.44\textwidth}
        \centering
        \includegraphics[width=\textwidth]{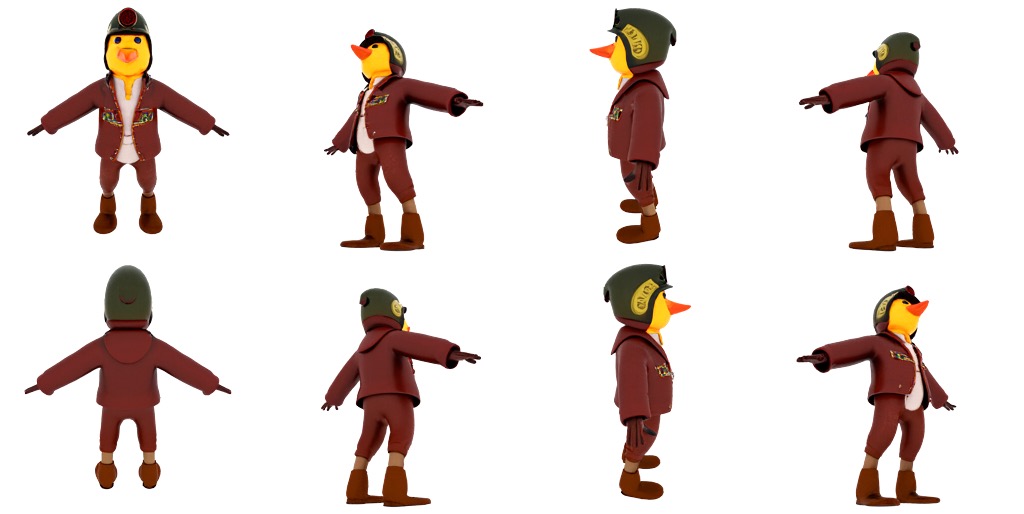}
    \end{subfigure}

    \centering
    Sampled 8 views

    \vspace{4pt}

    \begin{subfigure}[b]{0.46\textwidth}
        \centering
        \includegraphics[width=\textwidth]{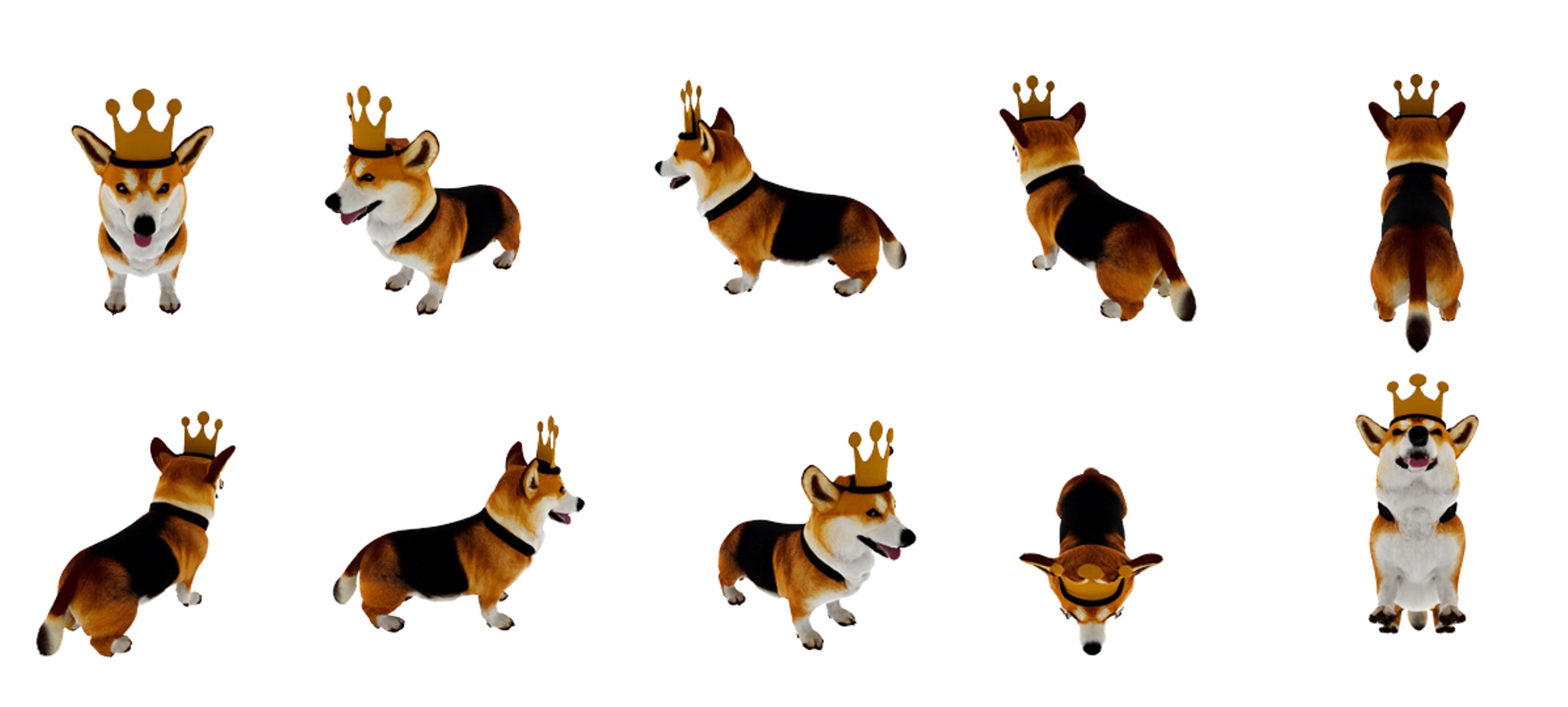}
    \end{subfigure}
    \hspace{24pt}
    \begin{subfigure}[b]{0.46\textwidth}
        \centering
        \includegraphics[width=\textwidth]{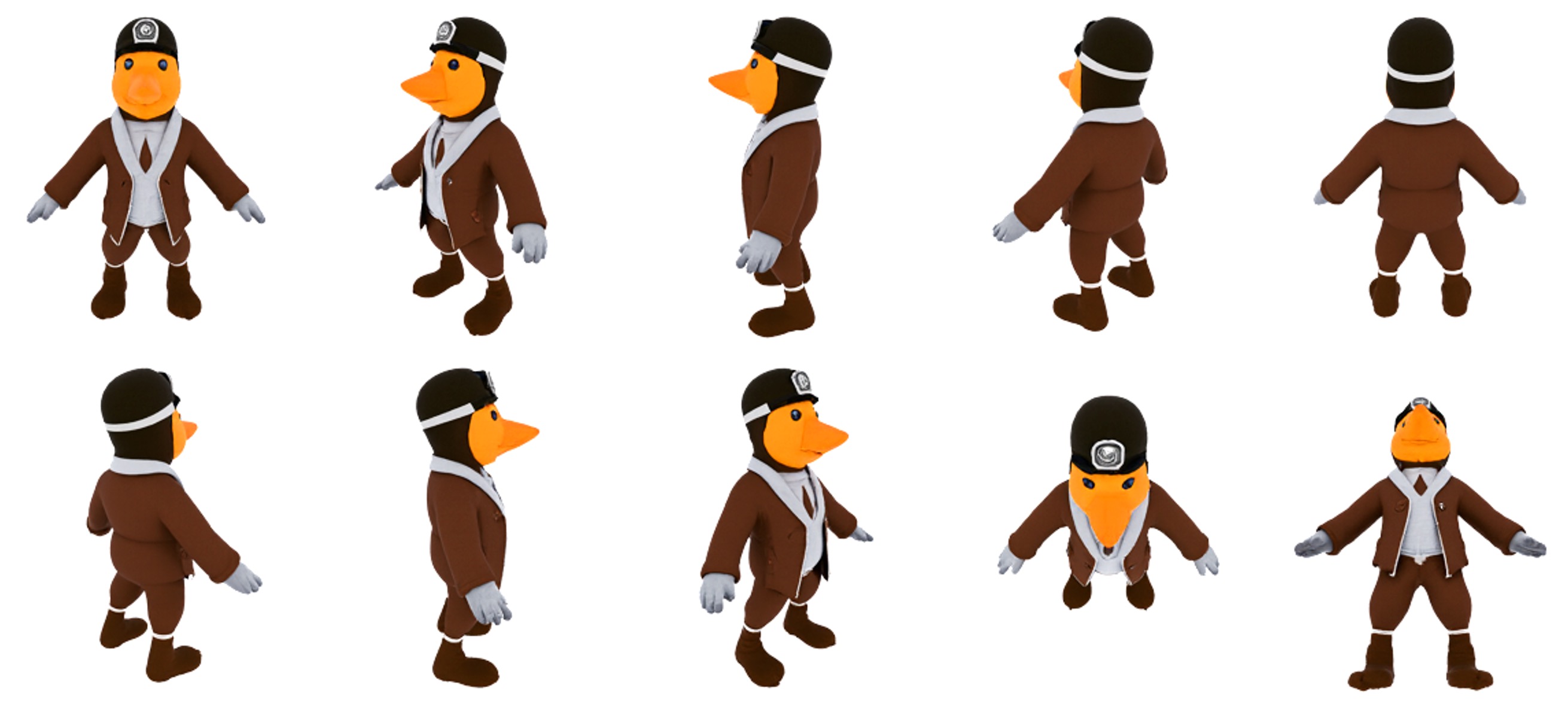}
    \end{subfigure}

    \centering
    Sampled 10 views

    \vspace{4pt}

    \begin{subfigure}[b]{0.48\textwidth}
        \centering
        \includegraphics[width=\textwidth]{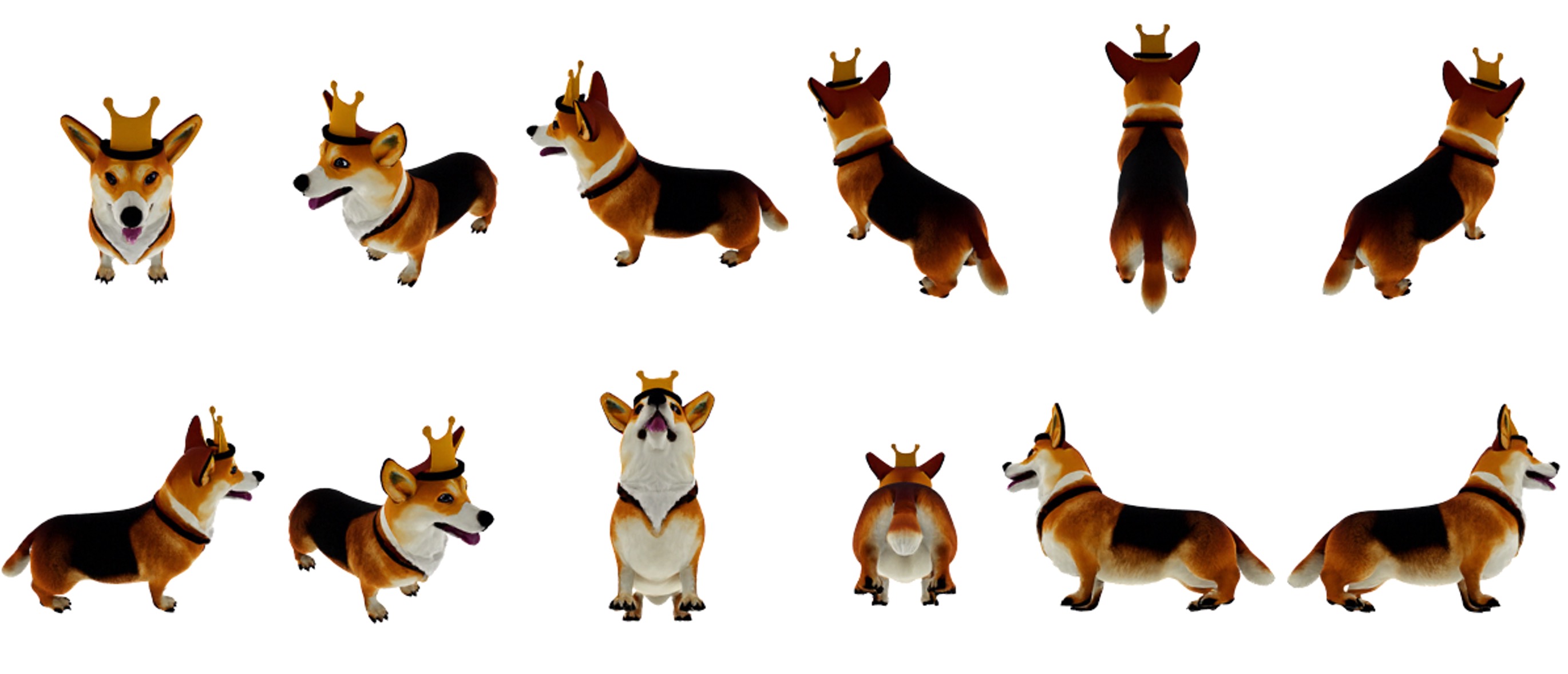}
    \end{subfigure}
    \hfill
    \begin{subfigure}[b]{0.48\textwidth}
        \centering
        \includegraphics[width=\textwidth]{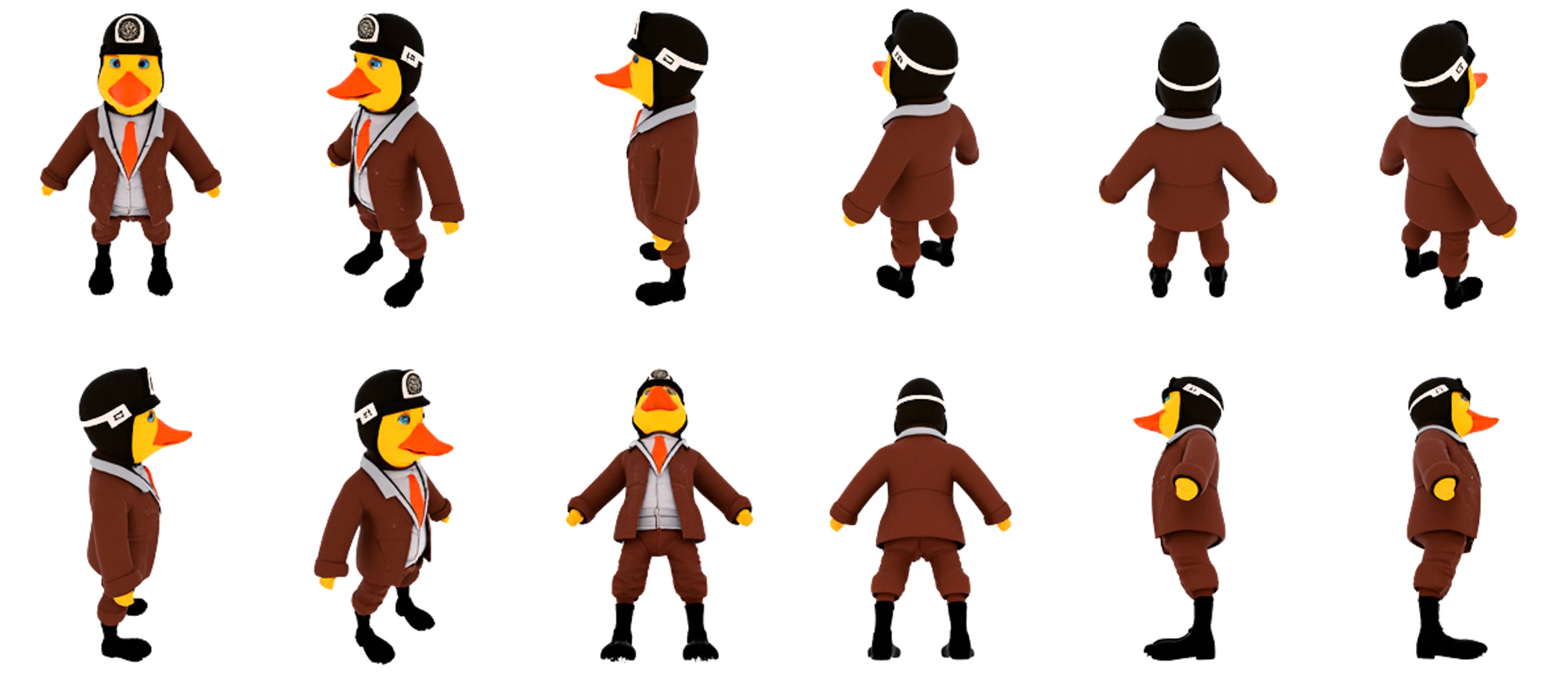}
    \end{subfigure}

    \centering
    Sampled 12 views

    \caption{
        \textbf{Comparison of number of sampled views.}
        All images are sampled from the same model.
        Our multi-view diffusion model can synthesize object images with dense viewpoint coverage while maintaining good multi-view consistency, making it suitable for the downstream reconstruction model.
    }
    \label{fig:mv_diffusion_sample_num_view}
\end{figure}

\subsection{Ablation Studies}
\vspace{4pt}
\noindent\textbf{Scaling with respect to the number of viewpoints.}
During inference, we can sample an arbitrary number of views while maintaining good multi-view consistency, as shown in~\cref{fig:mv_diffusion_sample_num_view}.
Generating more views allows for broader coverage of the object's regions in the multi-view images.
As we later discuss in~\cref{sec:recon}, the quality of the resulting 3D reconstruction is positively correlated to the number of multi-view observations~\citep{furukawa2009accurate}.
Therefore, the ability of the multi-view diffusion model to synthesize denser viewpoints is critical to the final 3D generation quality.

\begin{figure}[h!]
    \centering
    \small

    \begin{subfigure}[b]{0.48\textwidth}
        \centering
        \includegraphics[width=\textwidth]{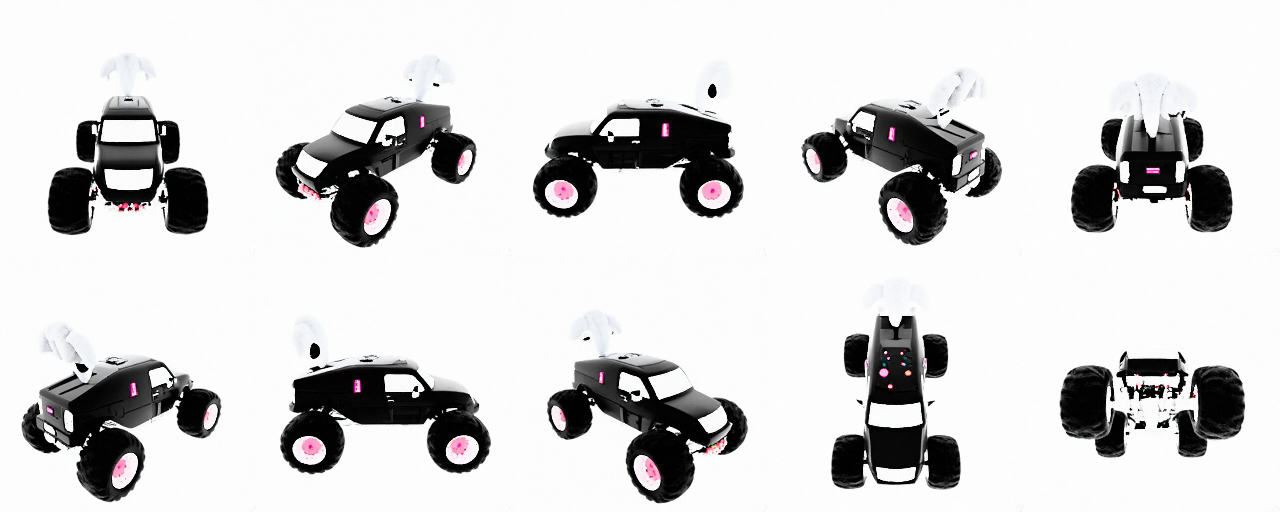}
    \end{subfigure}
    \hfill
    \begin{subfigure}[b]{0.48\textwidth}
        \centering
        \includegraphics[width=\textwidth]{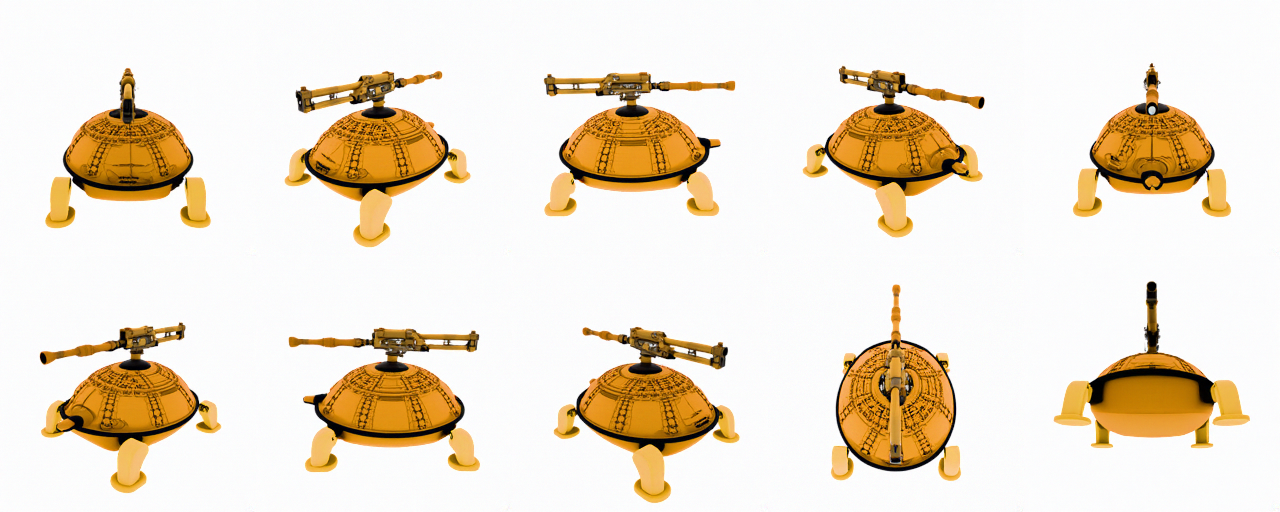}
    \end{subfigure}

    \centering
    Model trained with mostly 4 views

    \vspace{4pt}

    \begin{subfigure}[b]{0.48\textwidth}
        \centering
        \includegraphics[width=\textwidth]{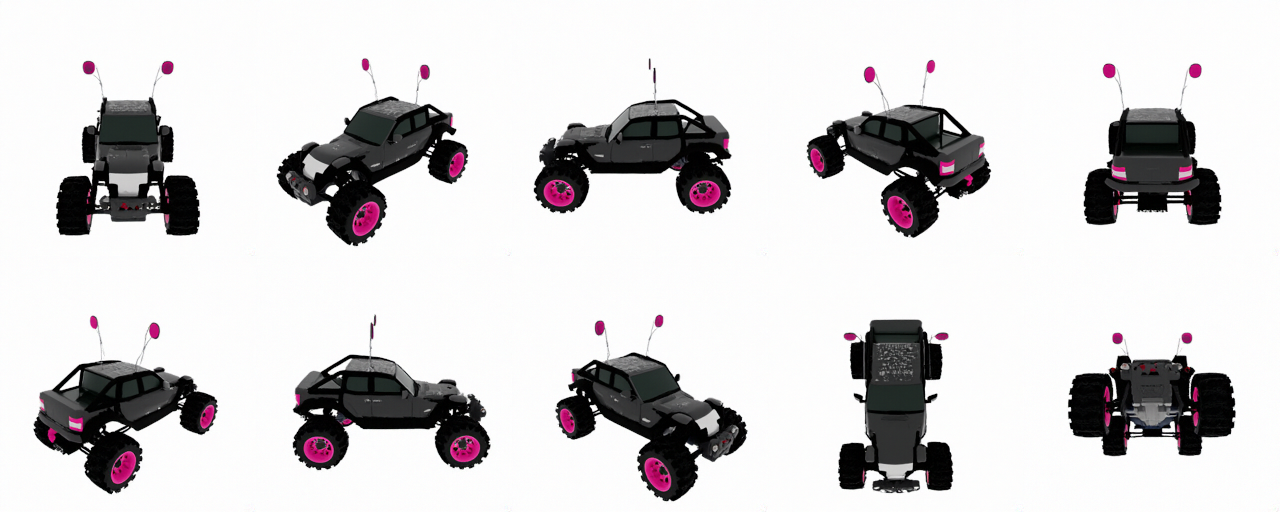}
    \end{subfigure}
    \hfill
    \begin{subfigure}[b]{0.48\textwidth}
        \centering
        \includegraphics[width=\textwidth]{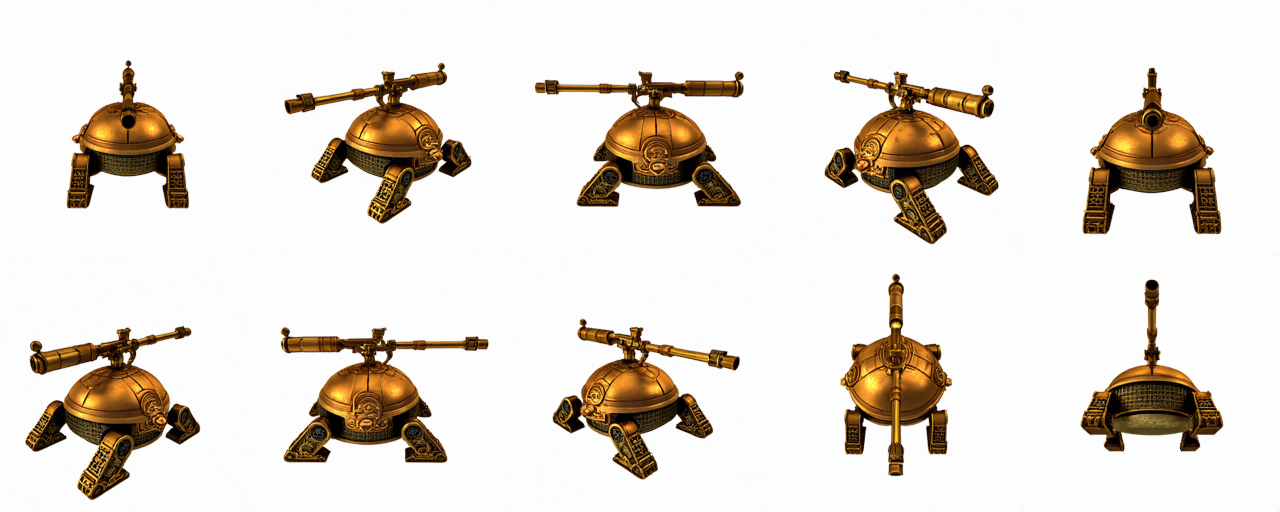}
    \end{subfigure}

    \centering
    Model trained with mostly 8 views

    \vspace{4pt}

    \caption{
        \textbf{Comparison of number of training views.}
        We compare two models trained primarily with different numbers of views (4 \vs 8), and sample images at the same 10 views at inference time.
        The model trained primarily on 8 views generates images with better multi-view consistency compared to the 4-view counterpart.
    }
    \label{fig:mv_diffusion_train_num_view}
\end{figure}

\vspace{4pt}
\noindent\textbf{Training across different numbers of viewpoints.}
During training, we sample 1, 4, or 8 views for each training object, assigning different sampling ratios for each number of views.
While training with a varying number of views enables sampling an arbitrary number of views during inference, it is still preferable to match the training views to those expected during inference.
This helps minimize the gap between training and inference performance.
We compare between two models -- one trained primarily on 4-view images and one on 8-view images -- and sample 10-view images at the same viewpoints.
As shown in~\cref{fig:mv_diffusion_train_num_view}, the model trained mostly with 8-view images produces more natural looking images with better multi-view consistency across views compared to that trained mostly with 4-view images.

\section{Reconstruction Model}
\label{sec:recon}

Extracting 3D structure from image observations is typically referred to as photogrammetry, which has been widely applied to many 3D reconstruction tasks~\citep{mildenhall2020nerf,wang2021neus,li2023neuralangelo}.
We use a Transformer-based reconstruction model~\citep{hong2023lrm} to generate the 3D mesh geometry, texture map, and material map from multi-view images.
We find that a Transformer-based model demonstrates strong generalization capabilities to unseen object images, including synthesized outputs from 2D multi-view diffusion models (\cref{sec:mvdiff}).

We use a decoder-only Transformer model with a latent 3D representation as triplanes~\citep{chan2023generative,hong2023lrm}.
The input RGB and normal images serve as conditioning for the reconstruction model, with cross-attention layers applied between triplane tokens and the input conditioning.
The triplane tokens are processed through MLPs to predict neural fields for Signed Distance Functions (SDF) and PBR properties~\citep{karis2013real}, which are used for SDF-based volume rendering~\citep{yariv2021volume}.
The neural SDF is converted into a 3D mesh through isosurface extraction~\citep{lorensen1998marching,shen2023flexible}.
The PBR properties are baked into texture and material maps via UV mapping, including albedo colors and material properties like roughness and metallic channels.

\vspace{4pt}

\noindent\textbf{Training.}
We train our reconstruction model using large-scale imagery and 3D asset data.
The model is supervised on depth, normal, mask, albedo, and material channels through SDF-based volume rendering, with outputs rendered from artist-generated meshes.
Since surface normal computation is relatively expensive, we compute the normal only at the surface and supervise against the ground truth.
We find that aligning the uncertainty of the SDF~\citep{yariv2021volume} with the corresponding rendering resolution improves the visual quality of the final output.
Additionally, we mask out object edges during loss computation to avoid noisy samples caused by aliasing.
To smooth noisy gradients across samples, we apply exponential moving average (EMA) to aggregate the final reconstruction model weights.

\noindent\textbf{Mesh post-processing.}
After obtaining the dense triangular 3D mesh from isosurface extraction, we post-process the mesh with the following steps:
\begin{enumerate}[leftmargin=18pt]
    \setlength\itemsep{0pt}
    \item Retopologize into a quadrilateral (quad) mesh with simplified geometry and adaptive topologies.
    \item Generate the UV mapping based on the resulting quad mesh topology.
    \item Bake the albedo and material neural fields into a texture map and material map, respectively.
\end{enumerate}
These post-processing steps make the resulting mesh more suitable for further editing, essential for artistic and design-oriented downstream applications.

\subsection{Ablation Studies}

We study the scaling properties of the reconstruction model in the following aspects: (a) the number of input images and (b) the tokens representing the shapes.

\begin{figure*}[!tb]
    \centering
    \includegraphics[width=0.9\linewidth]{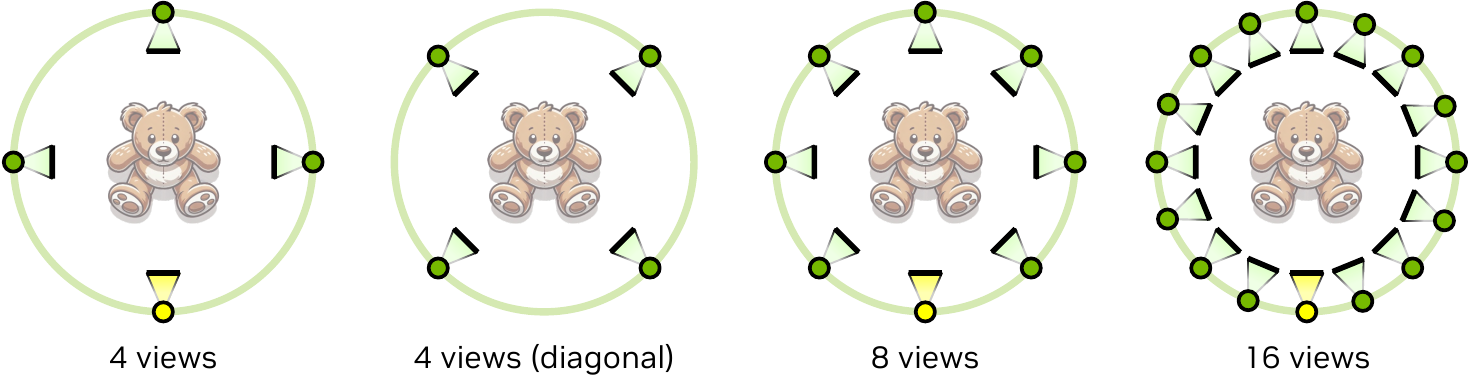}
    \caption{
        \textbf{Camera pose sets for scaling reconstruction.}
        The camera poses are set at various azimuth angles and a fixed elevation of $20\degree$.
        The camera in yellow indicates the ``frontal'' pose with respect to a canonicalized coordinate system.
        In each setup, the cameras are uniformly distributed around a circus at a fixed radius looking towards the origin.
        The ``4 views (diagonal)'' represents the same 4-view pose set but with a $45\degree$ offset, resulting in the cameras looking at the object region from diagonal angles.
    }
    \label{fig:camera}
\end{figure*}

\vspace{4pt}
\noindent\textbf{Experimental setup.}
For validation, we randomly select 78 shapes from a held-out dataset.
We report the LPIPS~\citep{zhang2018unreasonable} score on the albedo prediction to quantify the base texture reconstruction performance.
For material prediction accuracy, we use the $L_2$ error on the roughness and metallic values.
We also use the $L_2$ error between the ground-truth and predicted depths as a proxy for evaluating the geometry accuracy of the reconstructed shapes.
The camera poses used for input and output are fixed at an elevation of $20\degree$, pointing towards the origin (\cref{fig:camera}).

\vspace{4pt}
\noindent\textbf{Scaling with respect to the number of viewpoints.}
We find that our reconstruction model consistently recovers the input views more accurately than novel views.
The model scales well with respect to the number of viewpoints, \ie, its performance improves as more information is provided.
We demonstrate this with various combinations of input and output camera views, as shown in~\cref{fig:camera}.

\begin{table*}[!tb]
\setlength\tabcolsep{0.3em}
\centering

\noindent
\begin{minipage}{0.32\textwidth}
    \centering
    {\textbf{\scriptsize Albedo LPIPS error}\vspace{3pt}}
    \resizebox{\textwidth}{!}{%
        \renewcommand{\arraystretch}{1.4}
        \begin{tabular}{@{}c|cccc}
        \specialrule{.15em}{.1em}{.1em}
        Input & \multicolumn{4}{c}{Validation views} \\
        views & 4 & 4 (diag.) & 8 & 16 \\ \hline
        4 & \cellcolor[HTML]{e3f1cc} 0.0732 & 0.0791 & 0.0762  & 0.0768  \\
        4 (diag.) & 0.0802 & \cellcolor[HTML]{e3f1cc} 0.0756 & 0.0779  &  0.0783 \\
        8 & 0.0691 & 0.0698 & \cellcolor[HTML]{e3f1cc} 0.0695 & 0.0699 \\
        16 & 0.0687 & 0.0689 & 0.0688 & \cellcolor[HTML]{e3f1cc} 0.0687 \\
        \specialrule{.15em}{.1em}{.1em}
        \end{tabular}%
    }
\end{minipage}%
\hfill
\begin{minipage}{0.32\textwidth}
    \centering
    {\textbf{\scriptsize Material $L_2$ error}\vspace{3pt}}
    \resizebox{\textwidth}{!}{%
        \renewcommand{\arraystretch}{1.4}
        \begin{tabular}{@{}c|cccc}
        \specialrule{.15em}{.1em}{.1em}
        Input & \multicolumn{4}{c}{Validation views} \\
        views & 4 & 4 (diag.) & 8 & 16 \\ \hline
        4 & \cellcolor[HTML]{e3f1cc} 0.0015 & 0.0020 & 0.0017  & 0.0018  \\
        4 (diag.) & 0.0024 & \cellcolor[HTML]{e3f1cc} 0.0019 & 0.0022  &  0.0022 \\
        8 & 0.0013 & 0.0012 & \cellcolor[HTML]{e3f1cc} 0.0013 & 0.0013 \\
        16 & 0.0012 & 0.0013 & 0.0013 & \cellcolor[HTML]{e3f1cc} 0.0013 \\
        \specialrule{.15em}{.1em}{.1em}
        \end{tabular}%
    }
\end{minipage}%
\hfill
\begin{minipage}{0.32\textwidth}
    \centering
    {\textbf{\scriptsize Depth $L_2$ error}\vspace{3pt}}
    \resizebox{\textwidth}{!}{%
        \renewcommand{\arraystretch}{1.4}
        \begin{tabular}{@{}c|cccc}
        \specialrule{.15em}{.1em}{.1em}
        Input & \multicolumn{4}{c}{Validation views} \\
        views & 4 & 4 (diag.) & 8 & 16 \\ \hline
        4 & \cellcolor[HTML]{e3f1cc} 0.0689 & 0.0751 & 0.0720 & 0.0722 \\
        4 (diag.) & 0.0704 & \cellcolor[HTML]{e3f1cc} 0.0683 & 0.0694 &  0.0696 \\
        8 & 0.0626 & 0.0641 & \cellcolor[HTML]{e3f1cc} 0.0633 & 0.0633 \\
        16 & 0.0613 & 0.0626 & 0.0619 & \cellcolor[HTML]{e3f1cc} 0.0616 \\
        \specialrule{.15em}{.1em}{.1em}
        \end{tabular}%
    }
\end{minipage}%

\caption{
    \textbf{Comparison of the number of input views.}
    The diagonal cells indicate the cases where the input views match the validation views.
    The quality improves (error decreases) as the number of input views increases, demonstrating the scalability of our reconstruction model.
}
\label{tab:num_views}
\end{table*}

We present the quantitative results in~\cref{tab:num_views}.
The diagonal entries of the table are highlighted, as they represent cases where the inputs and outputs are identical.
These diagonal entries often show the best results in each row, indicating that the model reproduces the input views most accurately.
Additionally, the results consistently improve as the number of input views increases from 4 to 16.
This suggests that the reconstruction model benefits from the additional input information.

\begin{figure}[ht]
    \centering
    \begin{subfigure}{0.49\linewidth}
        \centering
        \includegraphics[width=\linewidth]{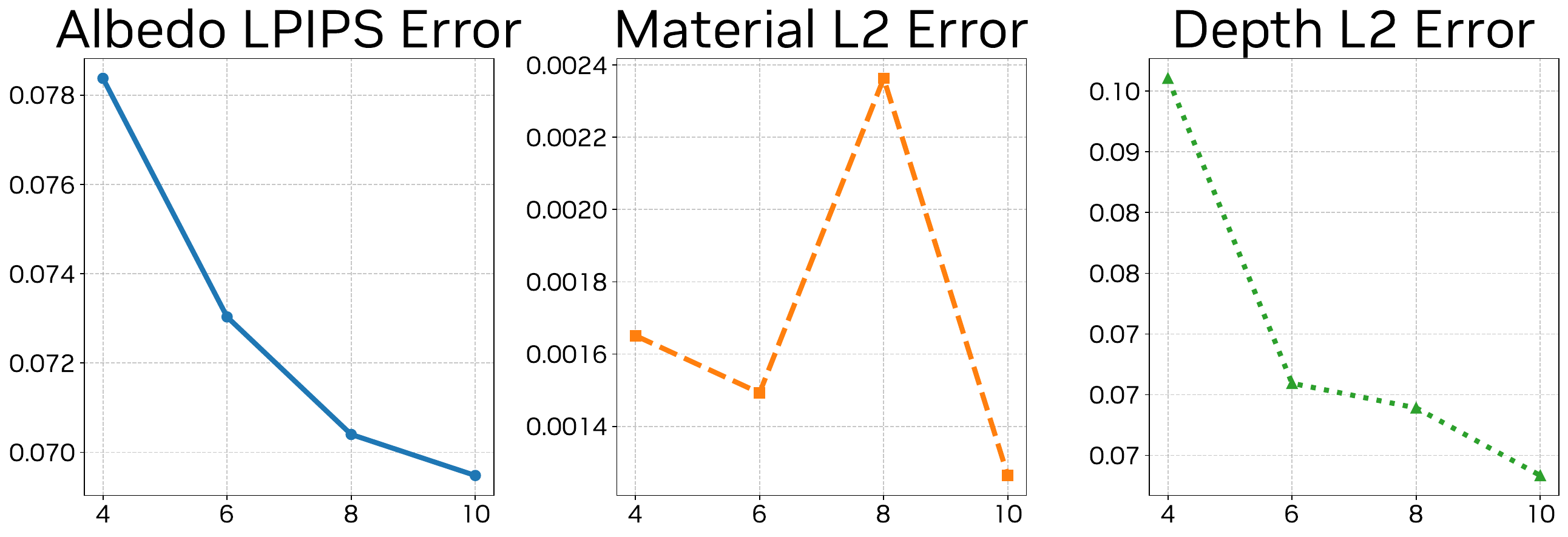}  %
        \caption{Number of Training Views}
        \label{fig:num_training_view}
    \end{subfigure}
    \hfill
    \begin{subfigure}{0.49\linewidth}
        \centering
        \includegraphics[width=\linewidth]{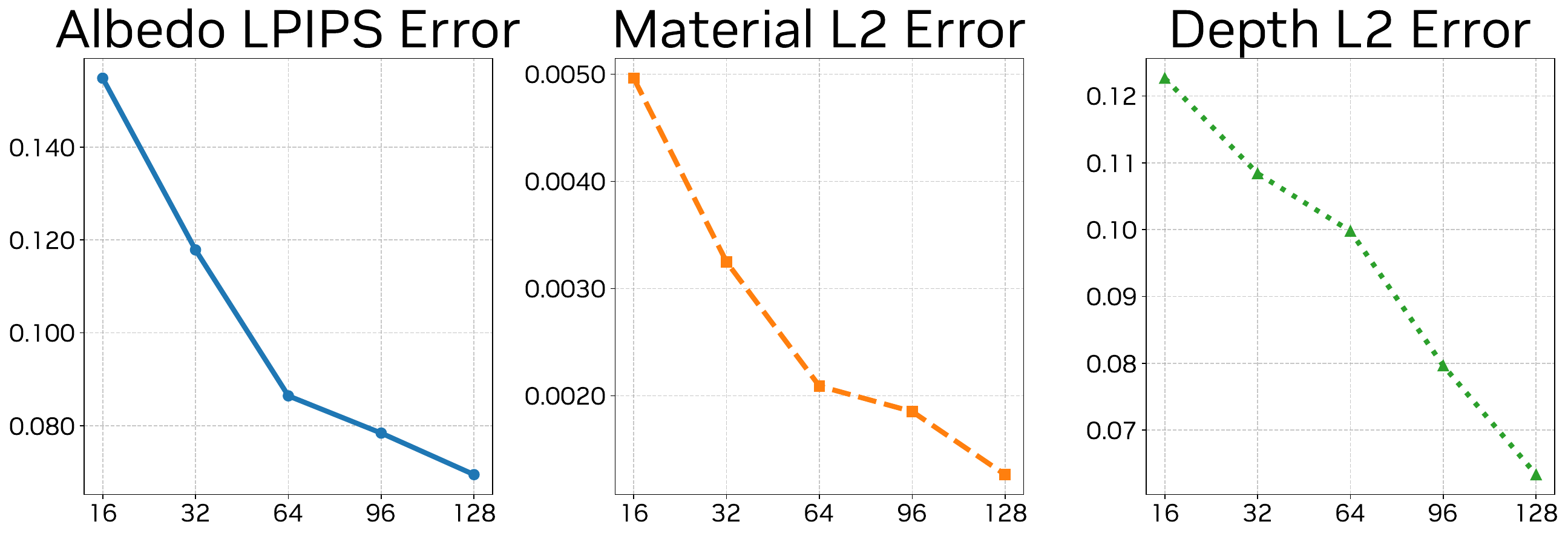}  %
        \caption{Triplane Resolution}
        \label{fig:num_tokens}
    \end{subfigure}

    \caption{
    (a) \textbf{Comparison of the number of training views.}
    The reconstruction model continues to improve as the number of training views increases.
    (b) \textbf{Comparison of the number of tokens.}
    With a fixed number of parameters, the model requires more compute with larger number of tokens. Note that the self-attention FLOPs increase quadratically with the triplane resolution. The quality consistently scales with the amount of compute.
    }
    \label{fig:num_view_num_tokens}
\end{figure}

Motivated by the model's scaling with the number of viewpoints, we further investigate whether the number of training views affects reconstruction quality.
We evaluate the model using a fixed 8-view setup, where the model is trained with 4, 6, 8, and 10 views.
The results are shown in~\cref{fig:num_training_view}.
Although stochastic sampling of camera poses provides diverse viewpoints during training, the reconstruction quality continues to improve as the number of training views in the same training step increases.

\captionsetup[subfigure]{justification=centering}

\vspace{4pt}
\noindent\textbf{Scaling with respect to compute.}
We study the impact of the compute requirements of the reconstruction model without changing the model size (\ie, the number of model parameters).
For this analysis, we scale down the triplane token sizes in the self-attention and cross-attention blocks to reduce computation.
Note that this adjustment does not alter the number of model parameters.
We observe from~\cref{fig:num_tokens} that as the number of tokens increases, the results improve proportionally with the available compute.

\section{Data Processing}
\label{sec:data}

Edify 3D is trained on a combination of non-public large-scale imagery, pre-rendered multi-view images, and 3D shape datasets.
We focus on pre-processing 3D shape data in this section.
The raw 3D data undergo several preprocessing steps to achieve the quality and format required for model training.

\vspace{4pt}
\noindent\textbf{Format conversion.}
The first step of the data processing pipeline involves converting all 3D shapes into a unified format.
We triangulate the meshes, pack all texture files, and convert the materials into metallic-roughness format.
We discard shapes with corrupted textures or materials.
This process results in a collection of 3D shapes that can be rendered as intended by the original creators.

\vspace{4pt}
\noindent\textbf{Quality filtering.}
We filter out non-object-centric data from the large-scale 3D datasets.
We render the shapes from multiple viewpoints and use AI classifiers to remove partial 3D scans, large scenes, shape collages, and shapes containing auxiliary structures such as backdrops and ground planes.
To ensure quality, this process is conducted through multiple rounds of active learning, with human experts continuously curating challenging examples to refine the AI classifier.
Additionally, we apply rule-based filtering to remove shapes with obvious issues, such as those that are excessively thin or lack texture.

\vspace{4pt}
\noindent\textbf{Canonical pose alignment.}
We align our training shapes to their canonical poses to reduce potential ambiguity when training the model.
Pose alignment is also achieved via active learning.
We manually curate a small number of examples, train a pose predictor, look for hard examples in the full dataset, and repeat the process.
Defining the canonical pose is also critical.
While many objects such as cars, animals, and shoes already have natural canonical poses, other shapes may lack a clear front side, in which case we define the functional part as the front and prioritize maintaining left-right symmetry.

\vspace{4pt}
\noindent\textbf{PBR rendering.}
To render the 3D data into images for the diffusion and reconstruction models, we use an in-house path tracer for photorealistic rendering.
We employ a diverse set of sampling techniques for the camera parameters.
Half of the images are rendered from fixed elevation angles with consistent intrinsic parameters, while the remaining images are rendered using random camera poses and intrinsics.
As a constraint, we maintain roughly consistent object sizes within the rendered images.
This approach accommodates both text-to-3D use cases (where there is full control over the camera parameters) and image-to-3D use cases (where the reference image may come from a wide range of camera intrinsics).

\vspace{4pt}
\noindent\textbf{AI captions.}
To caption the 3D shapes, we render one image per shape and use a vision-language model (VLM) to generate both long and short captions for the image.
To enhance the comprehensiveness of captions, we also provide the metadata of the shape (\eg title, description, category tree) to the VLM.

\begin{figure}[ht!]
    \centering
    \includegraphics[width=\linewidth]{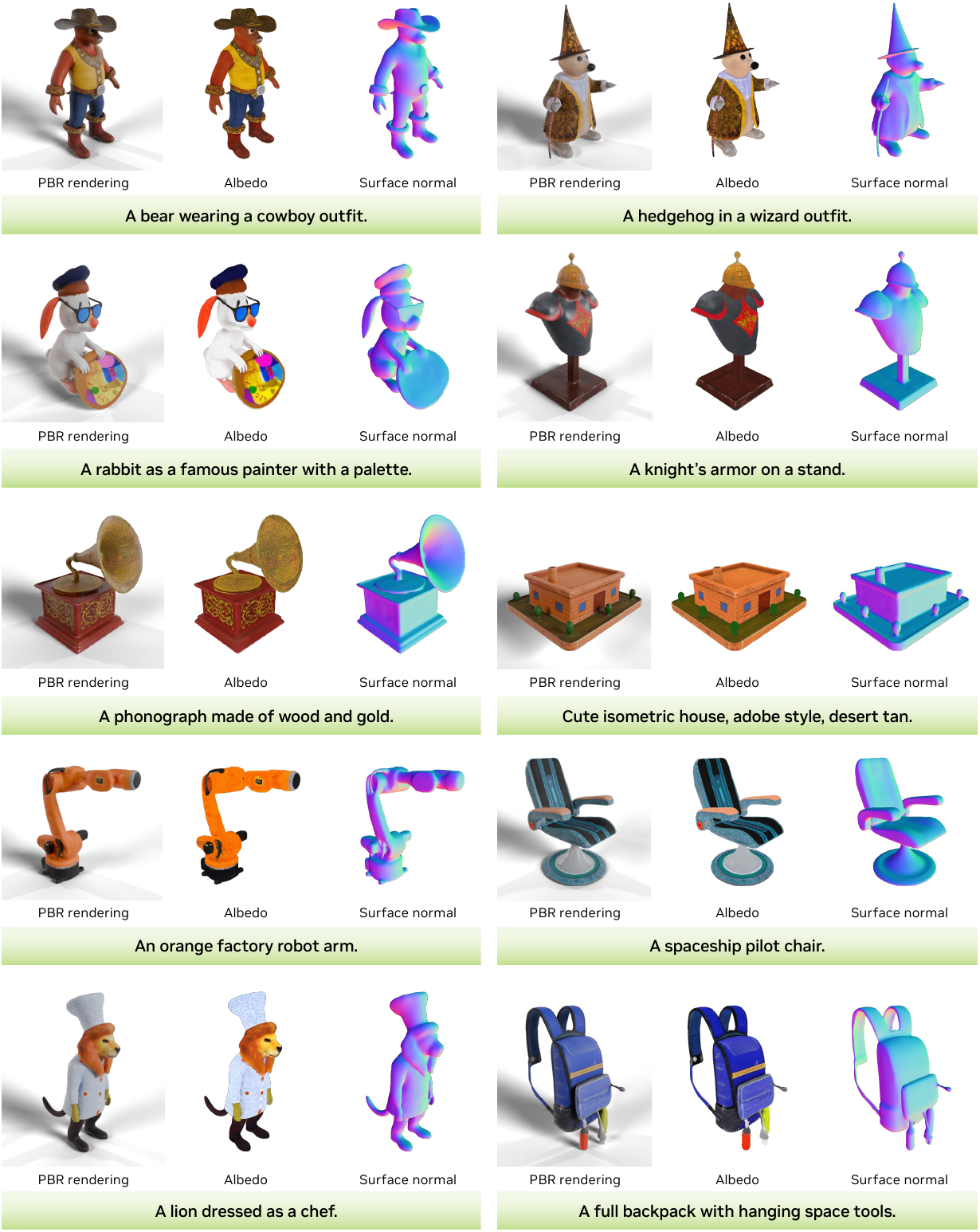}
    \caption{
        \textbf{Text-to-3D generation results.}
        We include the input text prompts as well as renderings and surface normals of generated assets.
        The generated 3D meshes include detailed geometry and sharp textures with well-decomposed albedo colors, making them suitable for various downstream editing and rendering applications.
    }
    \label{fig:text23d}
\end{figure}

\begin{figure}[t!]
    \centering
    \includegraphics[width=\linewidth]{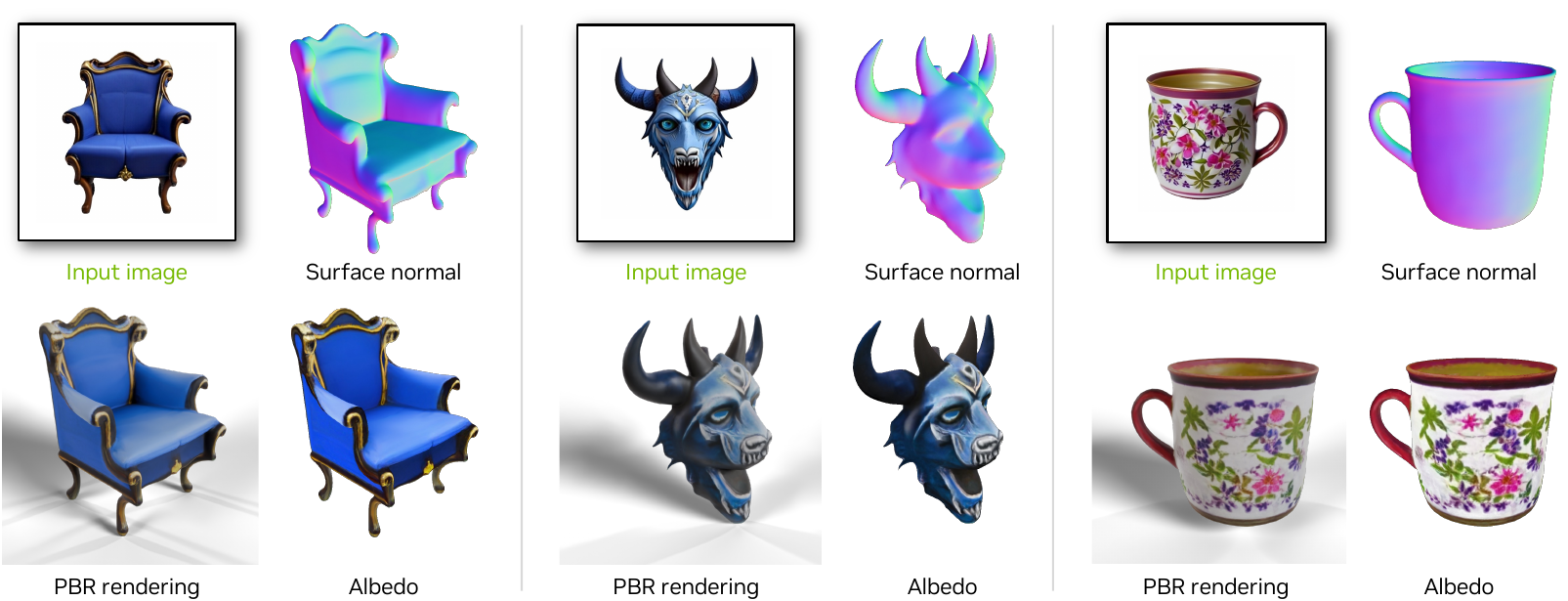}
    \caption{
        \textbf{Image-to-3D generation results.}
        We visualize the input reference images as well as renderings and surface normals of generated assets.
        Edify 3D can faithfully recover the underlying 3D structures of the reference object while also being able to hallucinate detailed textures in unseen surface regions (\eg, the backside of the cup).
    }
    \label{fig:image23d}
\end{figure}

\section{Results}
\label{sec:results}

We showcase text-to-3D generation results from Edify 3D in~\cref{fig:text23d} and image-to-3D generation in~\cref{fig:image23d}.
The generated meshes include detailed geometry and sharp textures, with well-decomposed albedo colors that represent the surface's base color.
For image-to-3D generation, Edify 3D not only accurately recovers the underlying 3D structures of the reference object, but it also can generate detailed textures in regions of the surface not directly observed in the input image.

\begin{figure}[t!]
    \centering
    \includegraphics[width=0.95\linewidth]{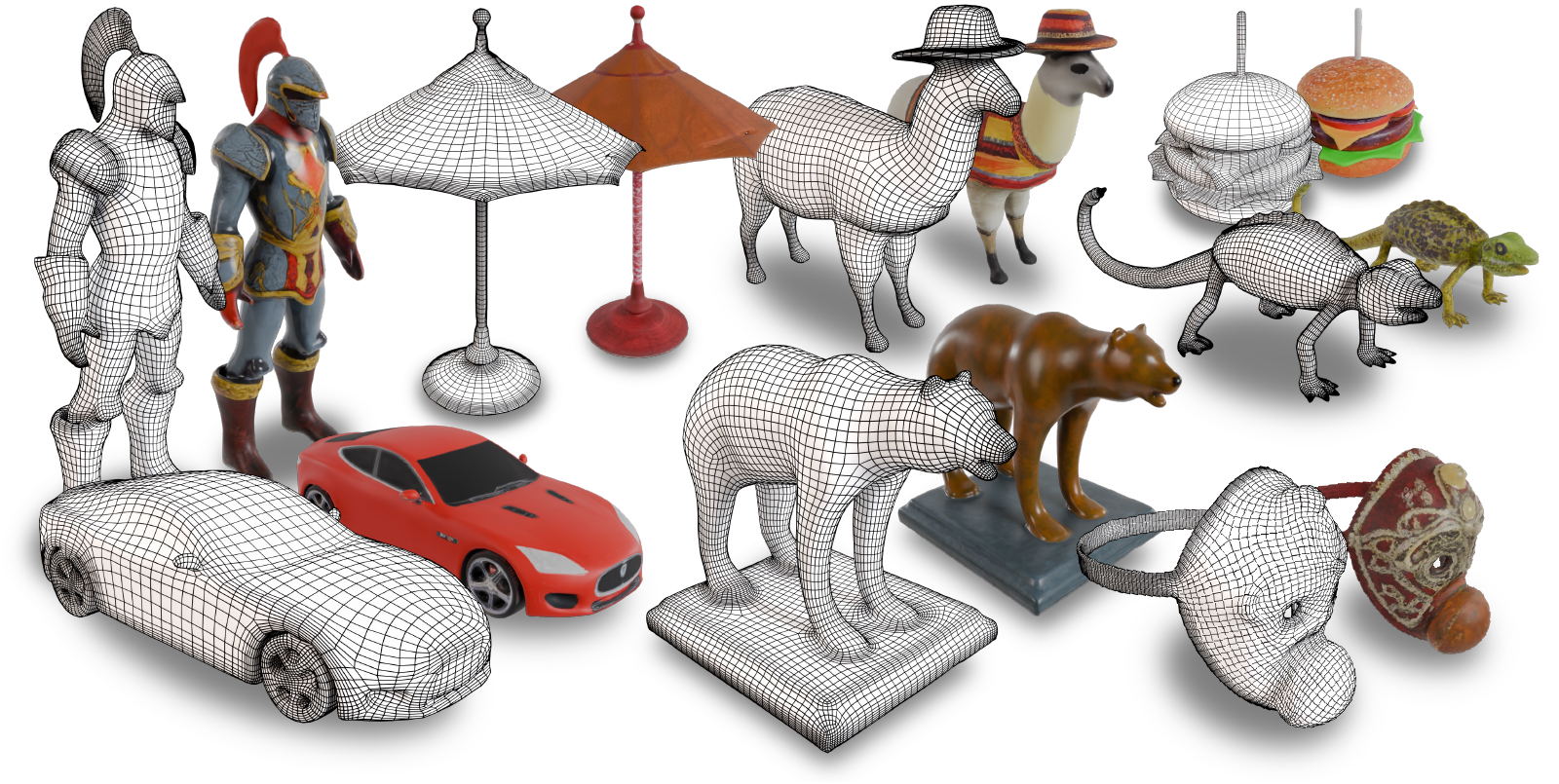}
    \caption{
        \textbf{Quad mesh topologies.}
        Edify 3D generates assets in the form of quad meshes with clean topologies, making it suitable for downstream editing workflows.
        We visualize the quad mesh topologies of the generated assets with their PBR renderings side-by-side.
    }
    \label{fig:quad}
\end{figure}

The assets generated by Edify 3D come in the form of quad meshes with well-organized topologies, as visualized in~\cref{fig:quad}.
These structured meshes allow for easier manipulation and precise adjustments, making them well-suited for various downstream editing tasks and rendering applications.
This enables seamless integration into 3D workflows that require visual fidelity and flexibility.

\section{Related Work}
\label{sec:related}

\noindent\textbf{3D asset generation.}
The challenge of 3D asset generation is often addressed by training models on 3D datasets~\citep{zeng2022lion,jun2023shap,nichol2022point,gupta20233dgen}, but the scarcity of these datasets limits the ability to generalize.
To overcome this, recent methods have shifted towards using models trained on large-scale image and video datasets.
Score Distillation Sampling (SDS)~\citep{poole2023dreamfusion} has been adopted in earlier methods~\citep{lin2023magic3d,wang2023score,wang2024prolificdreamer,zhu2023hifa,huang2023dreamtime,sun2023dreamcraft3d,yi2023gaussiandreamer,tang2023dreamgaussian} and extended to image-conditioned 3D generative models~\citep{liu2023zero,yu2023hifi,long2024wonder3d,tang2023make,qian2023magic123,wang2023imagedream}.
However, they often experience slow processing~\citep{lorraine2023att3d,xie2024latte3d} and are susceptible to issues such as the Janus face issue~\citep{shi2023mvdream}.
To improve performance, newer techniques integrate multi-view image generative models, focusing on producing multiple consistent views that can be reconstructed into 3D models~\citep{shi2023mvdream,shi2023zero123++,weng2023consistent123,liu2024one,yang2024consistnet,chan2023generative,tang2024mvdiffusion++,hollein2024viewdiff,gao2024cat3d,chen2024v3d}.
However, maintaining consistency across these views remains a challenge, leading to the development of methods that enhance reconstruction robustness from limited views~\citep{li2023instant3d,liu2024one++}.

\noindent\textbf{3D reconstruction from multi-view images.}
3D asset generation from limited views often involves 3D reconstruction techniques, often with differentiable rendering, that can utilize various 3D representations such as Neural Radiance Fields (NeRF)~\citep{mildenhall2020nerf}.
Meshes are the most commonly used format in industrial 3D engines, yet reconstructing high-quality meshes from multi-view images is challenging.
Traditional photogrammetry pipelines, including structure from motion (SfM)~\citep{agarwal2011building,snavely2006photo,schoenberger2016sfm}, multi-view stereo (MVS)~\citep{furukawa2009accurate,schonberger2016pixelwise}, and surface extraction~\citep{lorensen1998marching,kazhdan2006poisson,shen2021deep,shen2023flexible}, are costly and time-consuming, often yielding low-quality results.
While NeRF-based neural rendering methods can achieve high-quality 3D reconstructions~\citep{wang2021neus,yariv2021volume,li2023neuralangelo,kerbl20233d,guedon2024sugar,huang20242d}, they require dense images and extensive optimization, and converting radiance fields into meshes can lead to suboptimal results.
To address these limitations, Transformer-based~\citep{vaswani2017attention} models further improve 3D NeRF reconstruction from sparse views by learning a feed-forward prior~\citep{hong2023lrm}.

\noindent\textbf{Texture and material generation.}
Earlier approaches targeting 3D texture generation given a 3D shape include CLIP~\citep{radford2021learning} for text alignment~\citep{michel2022text2mesh,mohammad2022clip} and SDS loss optimization~\citep{poole2023dreamfusion}.
To improve 3D awareness, some text-to-3D methods combine texture inpainting with depth-conditioned diffusion~\citep{chen2023text2tex}, albeit slower and more artifact-prone.
To enhance consistency, other techniques alternate diffusion with reprojection~\citep{cao2023texfusion} or generate multiple textured views simultaneously~\citep{deng2024flashtex}, though at a higher computational cost.
To further enhance realism, some methods have enabled multi-view PBR modeling~\citep{zhang2021nerfactor,boss2022samurai} to extend support for generating material properties~\citep{chen2023fantasia3d,xu2023matlaber,qiu2024richdreamer}.

\section{Application: 3D Scene Generation}
\label{sec:scenegen}

In this section, we demonstrate an application of our Edify 3D model to scalable 3D scene generation~\citep{bahmani2023cc3d}.
High-quality, large-scale 3D scenes are pivotal for content creation and training robust embodied AI agents.
However, existing scene creation mostly depends on 3D scans with costly human annotations~\citep{dai2017scannet} or direct scene modeling~\citep{bautista2022gaudi,chen2023scenedreamer,fridman2024scenescape}, which largely limits the scalability.

\begin{figure}[t!]
    \centering
    \includegraphics[width=\linewidth,page=1]{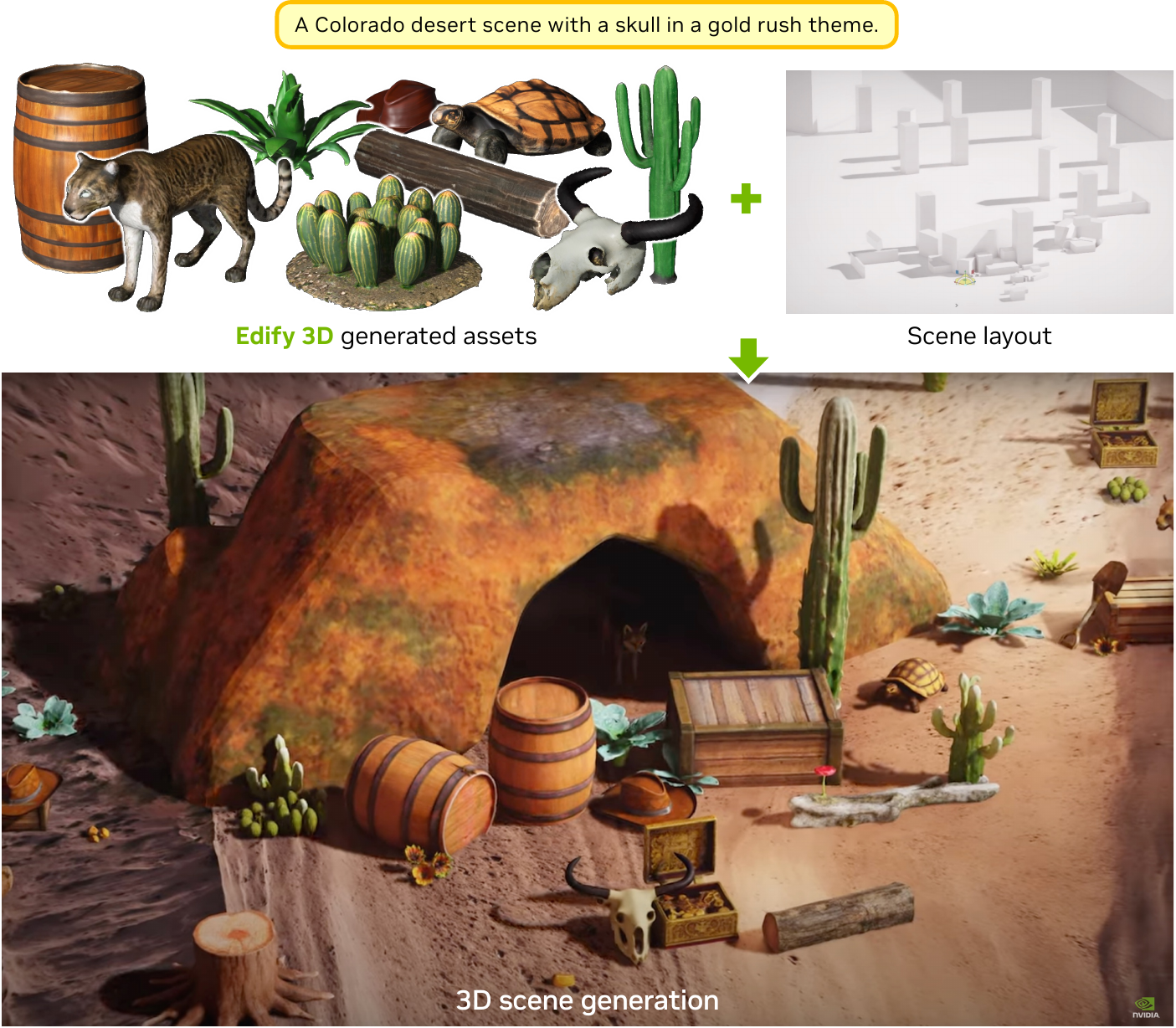}
    \caption{
        The high-quality 3D assets generated by Edify 3D can be extended to 3D scene generation combined with a given scene layout, which can be generated by an LLM.
        This enables the generated 3D scene to be readily rendered at high quality while being editable for various downstream applications.
    }
    \label{fig:scene-gen}
\end{figure}

With Edify 3D as the 3D asset generation API, we can design a scalable system to generate 3D scenes from only an input text prompt~\citep{lin2024genusd}.
The system generates the \emph{scene layout} of 3D objects with Large Language Models (LLM)~\citep{openai2023gpt}, which specifies the positions and sizes of 3D objects\footnote{We refer the readers to~\citet{lin2024genusd} for details on layout generation with LLMs.}.
As a result, the system can generate a realistic and complex 3D scene that coherently aligns with the text prompt describing the whole scene.

We show an example result of 3D scene generation in~\cref{fig:scene-gen}.
The generated assets from Edify 3D include detailed geometry and textures, forming a scene composition together with a generated scene layout.
Since all the 3D assets are created individually, the generated 3D scene is naturally editable for various specialized applications, such as artist creation, 3D designs, or embodied AI simulations.

\section{Conclusion}
\label{sec:conclusion}

In this technical report, we present Edify 3D, a solution designed for high-quality 3D asset generation.
We introduce the Edify 3D models, analyze the scaling laws of each model, and describe the data curation pipeline.
Additionally, we explore the application of Edify 3D to scalable 3D scene generation.
We are committed to advancing and developing new automation tools for 3D asset generation, making 3D content creation more accessible to everyone.

\appendix

\section{Contributors and Acknowledgements}
\label{sec:contributors}

\subsection{Core Contributors}

\textbf{System design:} Chen-Hsuan Lin, Xiaohui Zeng, Zhaoshuo Li, Zekun Hao, Ming-Yu Liu, Tsung-Yi Lin. \\
\textbf{Multi-view diffusion model:} Xiaohui Zeng, Qinsheng Zhang, Ming-Yu Liu, Tsung-Yi Lin. \\
\textbf{Reconstruction model:} Zhaoshuo Li, Chen-Hsuan Lin, Zekun Hao, Yen-Chen Lin, Ming-Yu Liu, Tsung-Yi Lin. \\
\textbf{3D data processing:} Zekun Hao, Fangyin Wei, Yin Cui, Yunhao Ge, Yifan Ding, Donglai Xiang, Qianli Ma, Jacob Munkberg, Jon Hasselgren, Chen-Hsuan Lin, Tsung-Yi Lin. \\
\textbf{Mesh post-processing:} Donglai Xiang, Qianli Ma, J.P. Lewis, Zekun Hao, Zhaoshuo Li, Fangyin Wei, Xiaohui Zeng, Jingyi Jin, Chen-Hsuan Lin, Tsung-Yi Lin. \\
\textbf{Evaluation:} Jingyi Jin, Xiaohui Zeng, Zhaoshuo Li, Qianli Ma, Yen-Chen Lin, Chen-Hsuan Lin, Tsung-Yi Lin.

\subsection{Contributors}
Maciej Bala, Jacob Huffman, Alice Luo, Stella Shi, Jiashu Xu.

\subsection{Acknowledgements}
We thank Lars Bishop, Sanja Fidler, Jun Gao, Jinwei Gu, Aaron Lefohn, Arun Mallya, Hanzi Mao, Seungjun Nah, Fitsum Reda, David Romero Guzman, Rohan Sawhney, Nicholas Sharp, Tianchang Shen, Peter Shipkov, Towaki Takikawa, Heng Wang, and Martin Watt for useful research discussions and prototyping.
We also thank Dane Aconfora, Yazdan Aghaghiri, Margaret Albrecht, Arslan Ali, Sivakumar Arayandi Thottakara, Amelia Barton, Lucas Brown, Matt Catrett, Douglas Chang, Steve Chappell, Gerardo Delgado Cabrera, John Dickinson, Amol Fasale, Daniela Flamm Jackson, Sandra Froehlich, Devika Ghaisas, Yugi Guvvala, Brett Hamilton, Mohammad Harrim, Nathan Horrocks, Akan Huang, Sophia Huang, Pooya Jannaty, Pranjali Joshi, Tobias Lasser, Gabriele Leone, Aaron Licata, Ashlee Martino-Tarr, Alexandre Milesi, Amanda Moran, Pawel Morkisz, Andrew Morse, Jashojit Mukherjee, Brad Nemire, Dade Orgeron, David Page, Mitesh Patel, Jason Paul, Joel Pennington, Lyne Tchapmi, Jibin Varghese, Thomas Volk, Raju Wagwani, Herb Woodruff, and Josh Young for feedback and support.

\clearpage
\setcitestyle{numbers}
\bibliographystyle{plainnat}
\bibliography{main}

\end{document}